\documentclass{article}
\usepackage[preprint]{neurips_2026}
\usepackage[utf8]{inputenc}
\usepackage[T1]{fontenc}
\usepackage{hyperref}
\usepackage{url}
\usepackage{booktabs}
\usepackage{amsfonts}
\usepackage{amsmath}
\usepackage{amssymb}
\usepackage{nicefrac}
\usepackage{microtype}
\usepackage{xcolor}
\usepackage{comment}
\usepackage{graphicx}

\title{A Graph-Based Inspection and Intervention Tool for \\
Assessing Mechanistic Learning in PINNs}

\author{%
Adwait Patkhedkar, Alifaraz Lakhani, Prathmesh Mohite, Abhijeet Salunke\\[2mm]
Sardar Patel Institute of Technology, Mumbai, India \\
Corresponding author :\texttt{adwait.patkhedkar23@spit.ac.in}
}

\begin{document}
\maketitle

\begin{abstract}
We ask whether physically meaningful correspondences discovered inside a
trained scientific model remain meaningful outside the conditions under
which they were discovered. We introduce GIIT (Graph-based Inspection and
Intervention Tool), which represents governing physics as a computational physics dependency graph,
maps graph nodes to internal network components via sensitivity- and
trend-based discovery, and tests the resulting mapping under targeted
intervention. Evaluating on temporal extrapolation out-of-distribution (OOD)
regimes of increasing severity, we find that temporal extrapolation is associated with layer-wise correspondence drift and reduced functional correspondence.
Specifically, while the model achieves low physics residual in-distribution
($1.915\times10^{-4}$ on full ID and $1.01\times10^{-4}$ on an ID sub-window $t \in [0.5, 0.8]$), the functional mapping changes even before leaving the training domain, and the shift continues as the evaluation window extends beyond the training domain: residual
error increases from $2.81\times10^{-2}$ on the boundary-crossing window ($t \in [0.5, 1.5]$) to $3.30\times10^{-1}$ on severe OOD ($t \in [1, 2]$),
accompanied by a systematic leftward shift of internal layer mappings, where
the average winner layer index drops from $5.71$ (ID) to $3.00$ (ID sub-window) down to $2.29$ (severe OOD).
Only $2$ of $7$ physical nodes maintain stable layer assignments across the boundary-crossing window,
and only $1$ of $7$ under severe extrapolation,
revealing potential internal functional correspondence changes before severe degradation in conventional output metrics.
Additional results for linear oscillatory systems are further detailed in the appendix.

\vspace{0.5em}
\end{abstract}

\section{Introduction}
\label{sec:intro}
Physics-Informed Neural Networks (PINNs) \citep{raissi2019pinn} embed
differential-equation residuals directly into the training loss, achieving
strong performance in low-data, high-simulation-cost regimes. However, a
near-zero physics residual and low output error are a form of
\emph{numerical} constraint satisfaction, not a guarantee of
\emph{mechanistic} correctness: a trained PINN's hidden layers may encode a
representation with little correspondence to the causal structure of the
system it was fit to \citep{wang2024understanding,krishnapriyan2021characterizing}.

General frameworks
for understanding model behavior \citep{murdoch2019pdr} and XAI approaches
tailored to scientific settings \citep{roscher2020xai} both emphasize that
explanations should align with domain knowledge, but the dominant tools
remain correlational: saliency and SHAP-style attribution, and
counterfactual explanation methods \citep{chou2022counterfactual}, operate
on model \emph{inputs} rather than internal components. Causal approaches
such as Causal Dependence Plots \citep{loftus2023cdp} and DAG-based SHAP
\citep{terminassian2023dag} use causal structure to sharpen input-level
explanations, but likewise stop at the input-output boundary. Work that
does look inside the model tends to go the other direction (discovering
causal structure \emph{from} data or representations
\citep{renero2025rex}), whereas we start from known physical relationships
and test whether a trained model is consistent with them. Operator-learning
methods such as DeepONet \citep{lu2021deeponet} extend PINN-style training
to function-space mappings but do not test internal correspondence either.
Our framing draws on the role of causal structure in scientific explanation
\citep{buijsman2023causal} and on causal abstraction as a theoretical basis
for mechanistic interpretability \citep{geiger2025mechinterp}, applying
these ideas to PINNs via an explicit mapping between physical variables and
internal tensors, followed by targeted intervention. Recent work has used PINNs to quantify the sensitivity of physical solutions to governing parameters through automatic differentiation. Existing PINN sensitivity analysis primarily studies physical-output sensitivity to physical parameters \citep{hanna2024sensitivity}; GIIT studies sensitivity and intervention at the level of internal neural representations.

This motivates our central question: \emph{do physically meaningful
correspondences discovered inside a model remain meaningful when the model
is evaluated outside the conditions under which the correspondence was
found?} Specifically, we evaluate whether internal functional correspondences remain consistent when transitioning from in-distribution interpolation
($t \in [0, 1]$) through progressively more severe temporal extrapolation regimes.

\textbf{Contributions.} We make three key contributions:
\begin{itemize}
    \item \textbf{The GIIT Pipeline:} We introduce a tool that automatically links physical concepts (like mass or damping) to specific internal layers of a neural network using sensitivity analysis.
    \item \textbf{Intervention Testing:} We provide a way to test these links. By applying controlled multiplicative perturbations to specific layers, we test whether altering the identified components changes the corresponding physical observables.
    \item \textbf{Out-of-Distribution (OOD) Analysis:} We test whether these intervention-validated functional correspondences hold up under distribution shift. By tracking mapping discovery across evaluation regimes, we observe a systematic shift toward earlier-layer winners as the evaluation window moves from full ID through late-ID, boundary-crossing, and severe OOD regimes.
\end{itemize}

\subsection{The GIIT Framework}
\begin{figure}[h]
\centering
\includegraphics[width=0.85\linewidth]{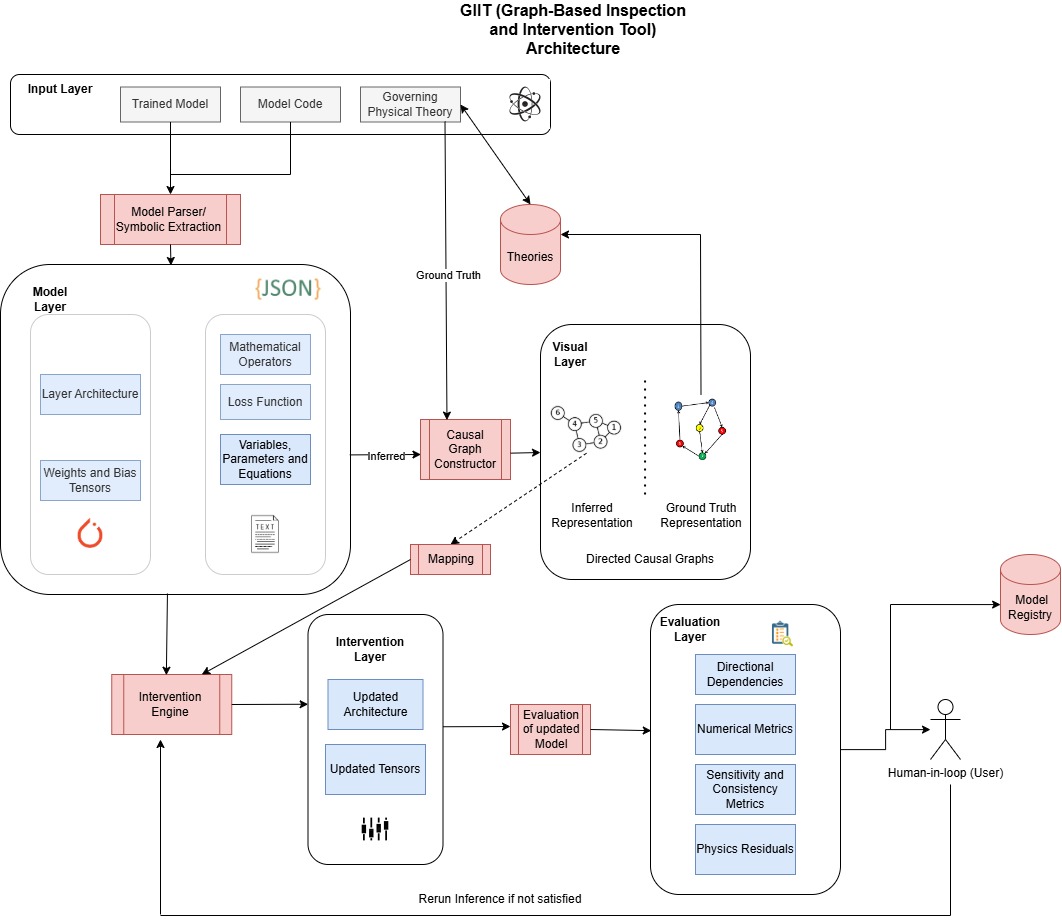}
\caption{GIIT pipeline: Physical specification
$\rightarrow$ Physics graph $\rightarrow$ Neural model $\rightarrow$
Variable/parameter mapping $\rightarrow$ Intervention $\rightarrow$ Physical
observables $\rightarrow$ ID$\to$OOD transfer test.}
\label{fig:overview}
\end{figure}

Figure~\ref{fig:overview} summarizes the pipeline: a physical specification
is compiled into a physics graph; a trained neural model is parsed for
candidate tensors; a variable/parameter mapping is discovered between graph
nodes and tensors; targeted interventions are applied to mapped tensors; and
the resulting change in physical observables is compared against the
analytically expected response.

\section{Method}
\label{sec:method}

\subsection{Physics Graph and Executable Ruleset}
We represent the governing physical mechanism as a directed graph
$G=(V,E)$, where $V$ is a set of physical variables (parameters, e.g. mass,
damping, viscosity; intermediate quantities, e.g. gradients; and outputs)
and $E$ encodes known directional dependencies (e.g. mass
$\rightarrow$ acceleration). Some nodes are direct model outputs; others are
derived via a small \emph{executable ruleset} on top of automatic
differentiation, e.g. $x \rightarrow x_t, x_{tt}$, or computed observables
such as amplitude $x \rightarrow A = \max(x) - \min(x)$. The graph can be
supplied directly as a JSON schema by a domain expert, or generated from
source code with optional LLM assistance. The graph specifies the physical
variables and their dependencies; the ruleset specifies \emph{how} derived
observables are computed from model output.

\subsection{Neural-Physical Mapping}
GIIT supports gradient-based sensitivity for output-derived variables and intervention-based sensitivity for parameters; in the experiments reported here, we use controlled perturbation sensitivity for the mapping analysis.

\textbf{Sensitivity-based gradient mapping.} For a node
corresponding to a quantity computable from model output via a known
formula $F_n$, GIIT can compute $q = \text{mean}(F_n(y))$ for $y=M(x)$ and evaluate gradient sensitivity across candidate layers $\ell$:
\begin{equation}
S_{\mathrm{grad}}(W_\ell, q) = \operatorname{mean}\left|\frac{\partial q}{\partial W_\ell}\right|.
\end{equation}

\textbf{Trend-based perturbation mapping.} Because governing parameters (e.g., mass, damping, or viscosity) are not direct network outputs, GIIT evaluates them through downstream behavioral observables computed by the ruleset (such as amplitude or frequency), assessing both response magnitude and directional consistency. Specifically, we apply a controlled multiplicative perturbation
$W_\ell' = (1+\epsilon)W_\ell$ (where the injected perturbation $\epsilon = \sigma|\mathcal{N}(0, 1)| \ge 0$ is strictly positive to preserve unambiguous directional trend evaluation), run a full forward pass, and measure the
relative change in the target observable,
\begin{equation}
S_{\mathrm{pert}} = \frac{|q(W_\ell') - q(W_\ell)|}{|q(W_\ell)| + \delta}.
\end{equation}
Candidate layers are ranked primarily by response magnitude; when a parameter has a known physical direction, we additionally evaluate whether the observed change has the expected sign. Tensor interventions are used to identify intervention-validated functional correspondences. Separately, analytical derivatives of the physics residual with respect to physical parameters ($\partial \mathcal{L}_R/\partial p$, where $\mathcal{L}_R = \operatorname{mean}(R^2)$) quantify residual-loss sensitivity.
Perturbing a tensor demonstrates that altering the identified component changes the target observable, rather than formal proof that the tensor uniquely or exclusively encodes the physical concept.

\subsection{Transfer Protocol}
\label{sec:transfer_protocol}
To test whether discovered functional mappings transfer beyond discovery conditions,
we define the full in-distribution training domain $\mathcal{D}_{\text{ID}} = \{t \in [0, 1]\}$,
an in-distribution sub-window $\mathcal{D}_{\text{sub}} = \{t \in [0.5, 0.8]\}$ (a subset of ID isolating late-time dynamics before the boundary),
and two progressively severe out-of-distribution extrapolation domains:
$\mathcal{D}_{\text{OOD}}^{(1)} = \{t \in [0.5, 1.5]\}$ (boundary-crossing window) and
$\mathcal{D}_{\text{OOD}}^{(2)} = \{t \in [1, 2]\}$ (severe extrapolation).
The protocol is:
\begin{enumerate}
    \item Discover the baseline node-to-layer mapping distribution $\mathcal{M}_{\text{ID}}: V \to \mathcal{L}$ across multiple perturbation replicates on one trained model using full ID data.
    \item Independently evaluate mapping distributions across the ID sub-window $\mathcal{D}_{\text{sub}}$ and each extrapolation domain $\mathcal{D}_{\text{OOD}}^{(i)}$ using the identical multi-replicate perturbation protocol.
    \item Quantify layer-wise correspondence drift by evaluating modal layer stability (whether the top-mapped component is preserved) and tracking systematic layer index shifts across regimes, alongside predictive error ($L_2$) and physics residual adherence ($\text{mean}(R^2)$).
\end{enumerate}
This tests whether functional correspondences remain stable under distribution shift, rather than relying only on output-level performance.

\section{Experimental Setup}
\label{sec:setup}
We evaluate GIIT across the four regimes defined in Section~\ref{sec:transfer_protocol}, using the 1D viscous Burgers' equation benchmark ($u_t + u u_x = \nu u_{xx}$) adapted from PINNsTorch \citep{bafghi2023pinnstorch} (\texttt{burgers\_continuous\_forward}), with spatial domain $x \in [-1, 1]$.
We evaluate predictive accuracy ($L_2$ error), physics adherence (mean residual norm), and internal functional correspondence using targeted interventions and parameter-gradient residual sensitivity.
For mapping discovery, we use perturbation sensitivity rather than gradient sensitivity because gradient sensitivity can also reflect bias parameters.
$L_2$ error and residuals are computed on a uniform $100 \times 100$ space-time grid against a high-resolution finite-difference reference.
For the Burgers' experiments, mapping analysis uses 100 perturbation replicates on a single trained model ($\sigma=0.1$, replicate index $\in [0, 99]$); the damped harmonic oscillator benchmark uses 21 perturbation replicates as detailed in Appendix~\ref{sec:appendix_dho}.

\section{Results}
\label{sec:results}

\subsection{Baseline Model Behavior}
In-distribution, the PINN achieves tight physics adherence and low prediction error (Table~\ref{tab:main_burgers}). Baseline perturbation mapping identifies the deepest layers as dominant: viscosity ($\nu$) and velocity ($u$) primarily map to \texttt{fch.5.0.weight}, while gradient and curvature terms map to \texttt{fch.6.0.weight}. However, viscosity shows an important failure: although consistently localized to \texttt{fch.5.0.weight}, its intervention response usually has the opposite of the expected physical direction. Thus, high sensitivity alone does not establish a physically valid correspondence.

\begin{table}[h]
\caption{Viscous Burgers' equation: ID (full and sub-window), boundary-crossing, and OOD extrapolation windows ($t \in [1, 2]$). A node mapping is defined as stable if its modal winning layer matches the full-ID baseline.}
\label{tab:main_burgers}
\centering
\footnotesize
\begin{tabular}{lcccc}
\toprule
Regime & $L_2$ (\%) & Residual $\text{mean}(R^2)$ & $\partial \mathcal{L}_R/\partial \nu$ & Stable Mappings \\
\midrule
ID ($t \in [0, 1]$)                          & $3.855$  & $1.915\times10^{-4}$ & $-1.705$   & --- (baseline) \\
ID Sub-window ($t \in [0.5, 0.8]$)          & $5.413$  & $1.011\times10^{-4}$ & $-5.064$   & 2/7 (28.6\%) \\
Boundary-crossing ($t \in [0.5, 1.5]$)      & $5.883$  & $2.811\times10^{-2}$ & $-73.576$  & 2/7 (28.6\%) \\
OOD ($t \in [1, 2]$)                        & $31.752$ & $3.301\times10^{-1}$ & $-154.115$ & 1/7 (14.3\%) \\
\bottomrule
\end{tabular}
\end{table}

\subsection{Correspondence Drift under Distribution Shift}
\begin{figure}[htbp]
\centering
\begin{minipage}{\linewidth}
  \centering
  \includegraphics[width=0.6\linewidth]{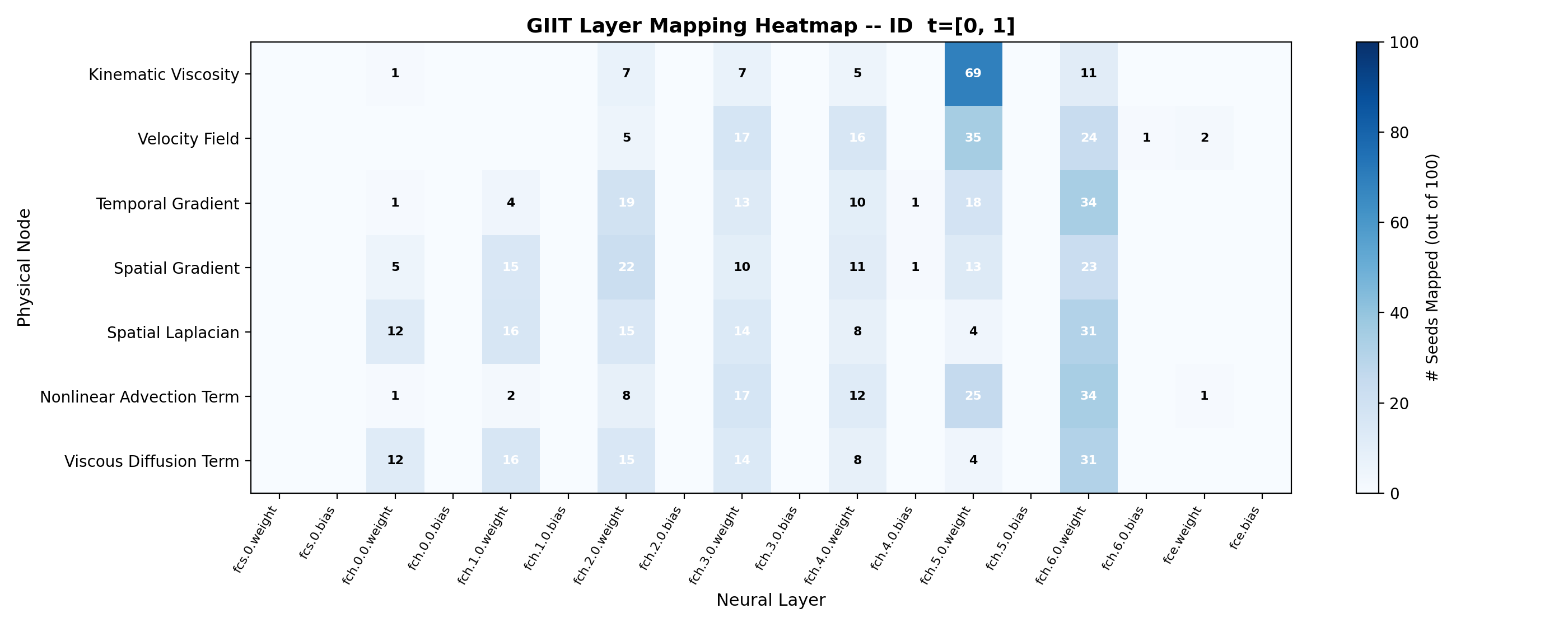}
  \vspace{0.1em}
  \centerline{\small \textbf{(a)} In-Distribution ($t \in [0, 1]$)}
\end{minipage}
\vspace{0.5em}

\begin{minipage}{\linewidth}
  \centering
  \includegraphics[width=0.6\linewidth]{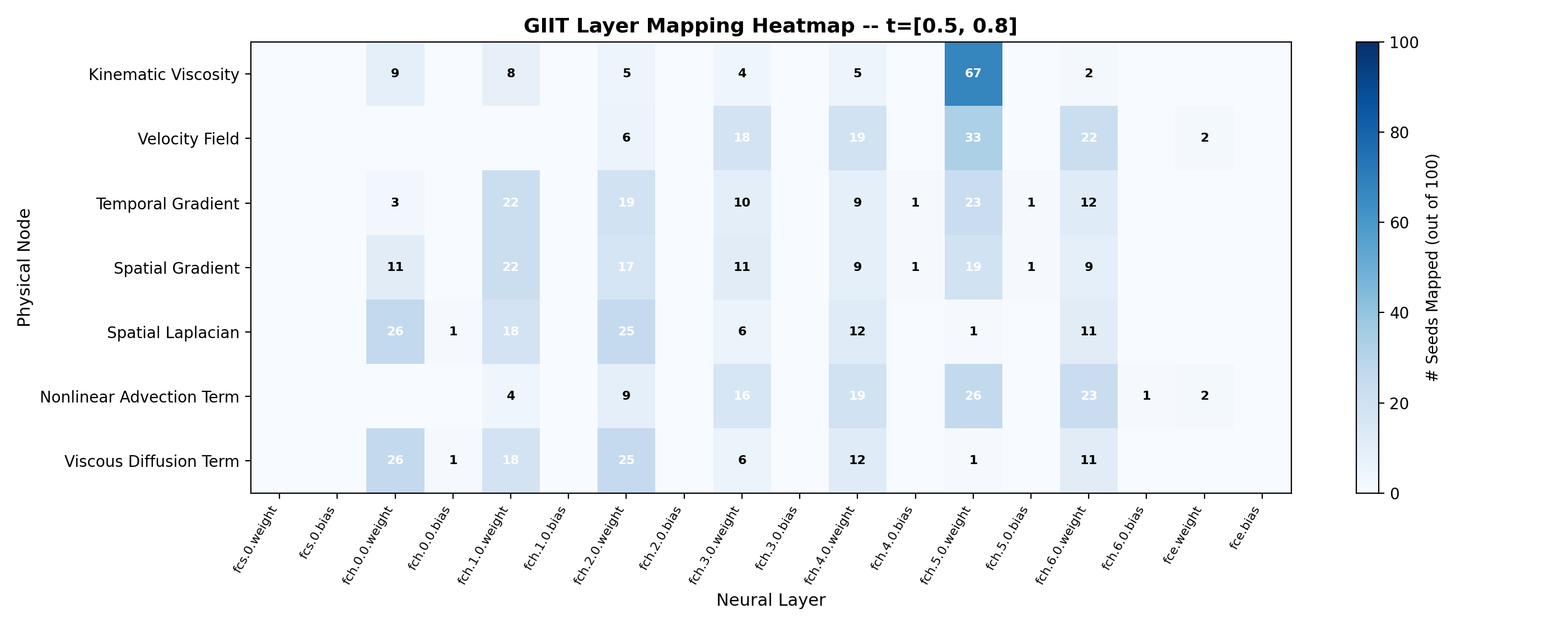}
  \vspace{0.1em}
  \centerline{\small \textbf{(b)} ID Sub-window ($t \in [0.5, 0.8]$)}
\end{minipage}
\vspace{0.5em}

\begin{minipage}{\linewidth}
  \centering
  \includegraphics[width=0.6\linewidth]{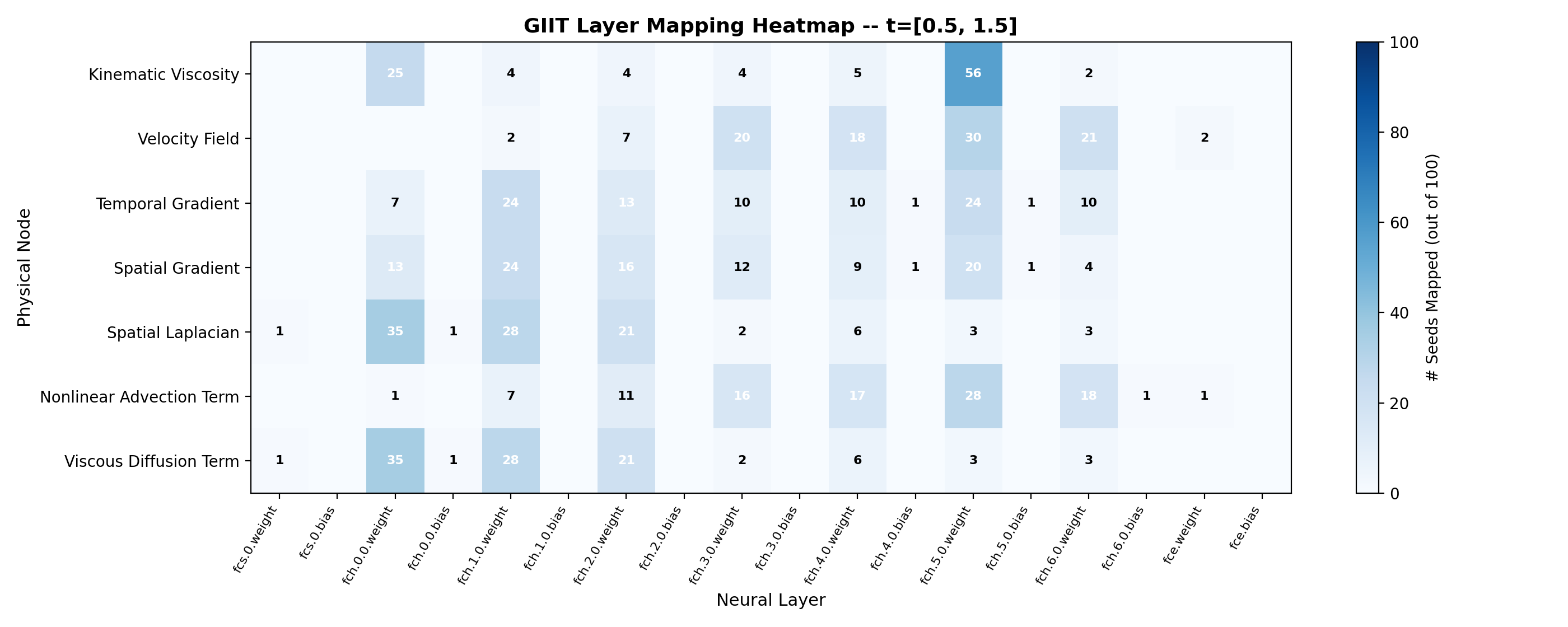}
  \vspace{0.1em}
  \centerline{\small \textbf{(c)} Boundary-crossing window ($t \in [0.5, 1.5]$)}
\end{minipage}
\vspace{0.5em}

\begin{minipage}{\linewidth}
  \centering
  \includegraphics[width=0.6\linewidth]{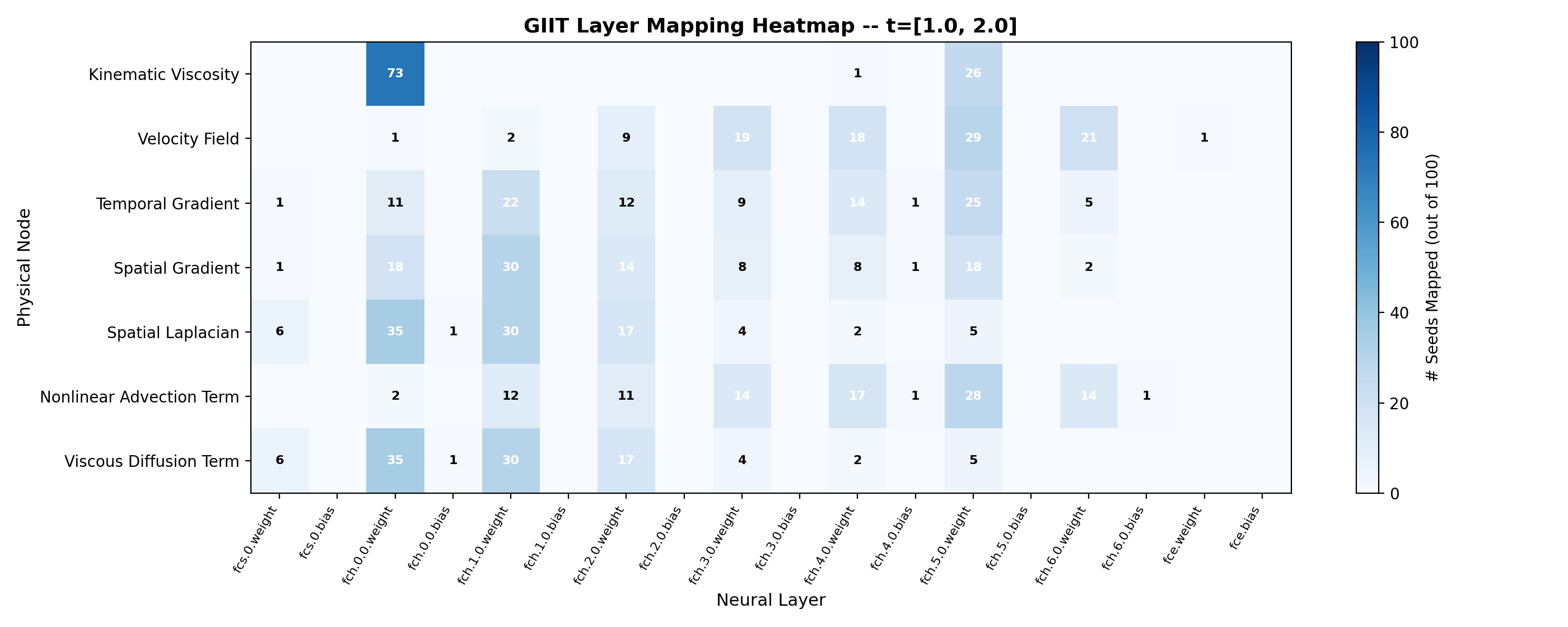}
  \vspace{0.1em}
  \centerline{\small \textbf{(d)} OOD Extrapolation ($t \in [1, 2]$)}
\end{minipage}
\caption{GIIT layer mapping distributions across 100 perturbation replicates ($\sigma = 0.1$) for Burgers' equation under ID and OOD regimes. Cell values show winner counts per layer, highlighting systematic leftward migration ($5.71 \to 2.29$).}
\label{fig:burgers_heatmaps}
\end{figure}

Mapping distributions shift toward earlier layers as evaluation moves away from the full-ID regime (Table~\ref{tab:burgers_seed_sweep}, Figure~\ref{fig:burgers_heatmaps}). This reorganization is already evident within the training domain: curvature-related terms ($u_{xx}, \nu u_{xx}$) migrate to the earliest hidden layer (\texttt{fch.0.0.weight}) in the late-time ID sub-window, reducing stable mappings to 2/7. The drift continues through boundary-crossing into OOD extrapolation, where only velocity ($u$) retains its baseline modal assignment.

\begin{table}[h]
\caption{Neural Component Mapping Distributions across 100 Perturbation Replicates on One Trained Model ($\sigma = 0.1$, replicates $\in [0, 99]$) for ID (full and sub-window) and extrapolation regimes. The average winner layer index (bottom row) is the mean modal hidden layer index ($k \in [0, 6]$ for \texttt{h$k$w}) across all 7 physical nodes. Abbreviations: \texttt{h$k$w} $=$ \texttt{fch.$k$.0.weight}.}
\label{tab:burgers_seed_sweep}
\centering
\footnotesize
\begin{tabular}{lcccccccc}
\toprule
Physical Node & \multicolumn{2}{c}{ID ($t \in [0, 1]$)} & \multicolumn{2}{c}{ID Sub ($[0.5, 0.8]$)} & \multicolumn{2}{c}{Bound.-cross ($[0.5, 1.5]$)} & \multicolumn{2}{c}{OOD ($[1, 2]$)} \\
\cmidrule(lr){2-3} \cmidrule(lr){4-5} \cmidrule(lr){6-7} \cmidrule(lr){8-9}
 & Top & Cons. & Top & Cons. & Top & Cons. & Top & Cons. \\
\midrule
Kinematic Viscosity ($\nu$)      & \texttt{h5w} & 69/100 & \texttt{h5w} & 67/100 & \texttt{h5w} & 56/100 & \texttt{h0w} & 73/100 \\
Velocity Field ($u$)             & \texttt{h5w} & 35/100 & \texttt{h5w} & 33/100 & \texttt{h5w} & 30/100 & \texttt{h5w} & 29/100 \\
Temporal Gradient ($u_t$)        & \texttt{h6w} & 34/100 & \texttt{h5w} & 23/100 & \texttt{h1w} & 24/100 & \texttt{h5w} & 25/100 \\
Spatial Gradient ($u_x$)         & \texttt{h6w} & 23/100 & \texttt{h1w} & 22/100 & \texttt{h1w} & 24/100 & \texttt{h1w} & 30/100 \\
Spatial Laplacian ($u_{xx}$)     & \texttt{h6w} & 31/100 & \texttt{h0w} & 26/100 & \texttt{h0w} & 35/100 & \texttt{h0w} & 35/100 \\
Nonlinear Advection ($u u_x$)    & \texttt{h6w} & 34/100 & \texttt{h5w} & 26/100 & \texttt{h5w} & 28/100 & \texttt{h5w} & 28/100 \\
Viscous Diffusion ($\nu u_{xx}$) & \texttt{h6w} & 31/100 & \texttt{h0w} & 26/100 & \texttt{h0w} & 35/100 & \texttt{h0w} & 35/100 \\
\midrule
\textbf{Avg.\ layer index}       & \multicolumn{2}{c}{$5.71$} & \multicolumn{2}{c}{$3.00$} & \multicolumn{2}{c}{$2.43$} & \multicolumn{2}{c}{$2.29$} \\
\bottomrule
\end{tabular}
\end{table}

\subsection{Output Degradation vs.\ Internal Correspondence}
Output-level performance and internal representation stability diverge across regimes (Table~\ref{tab:main_burgers}). The boundary-crossing regime reveals substantial internal reorganization despite relatively stable predictive error ($L_2 \approx 5.88\%$). In the OOD regime ($t \in [1, 2]$), predictive accuracy deteriorates sharply, accompanied by a corresponding increase in physics residual. Residual-loss sensitivity also increases under distribution shift, remaining negative as increasing $\nu$ locally reduces the physics residual. However, it does not capture the layer-wise correspondence shifts detected by GIIT.

\section{Discussion and Limitations}
\label{sec:limitations}
The observed leftward migration suggests that the functional role of network layers is context-dependent. One possibility is that deeper representations become less sensitive during extrapolation, causing interventions to select earlier layers; activation-level analysis is needed to distinguish this from a broader redistribution of sensitivity. Thus, internal representation stability cannot be inferred from low output loss alone.

The identified tensors should be interpreted as intervention-validated functional correspondences, not unique causal encodings or isolated modular circuits. The viscosity result further shows that high response magnitude does not guarantee the expected physical direction. Finally, our evidence is limited to two low-dimensional dynamical systems, so generalization to larger or higher-dimensional SciML models remains untested.

\section{Conclusion}
\label{sec:conclusion}
GIIT provides a model-agnostic framework for testing whether internal neural representations retain functional correspondence with known physical structure. Our experiments show that these correspondences can change substantially under temporal distribution shift, including within the nominal training domain. This suggests that output-level accuracy and physics residuals alone may be insufficient for diagnosing changes in internal functional organization. Future work should test whether these effects persist across independently trained models and higher-dimensional scientific systems.
\newpage
\bibliographystyle{plainnat}
\bibliography{references}

\newpage
\appendix
\section{Technical Appendix: Viscous Burgers' PINN Architecture and Trajectory Analysis}
\label{sec:appendix_burgers_base}
\begin{figure}[!htbp]
\centering
\includegraphics[width=0.92\linewidth]{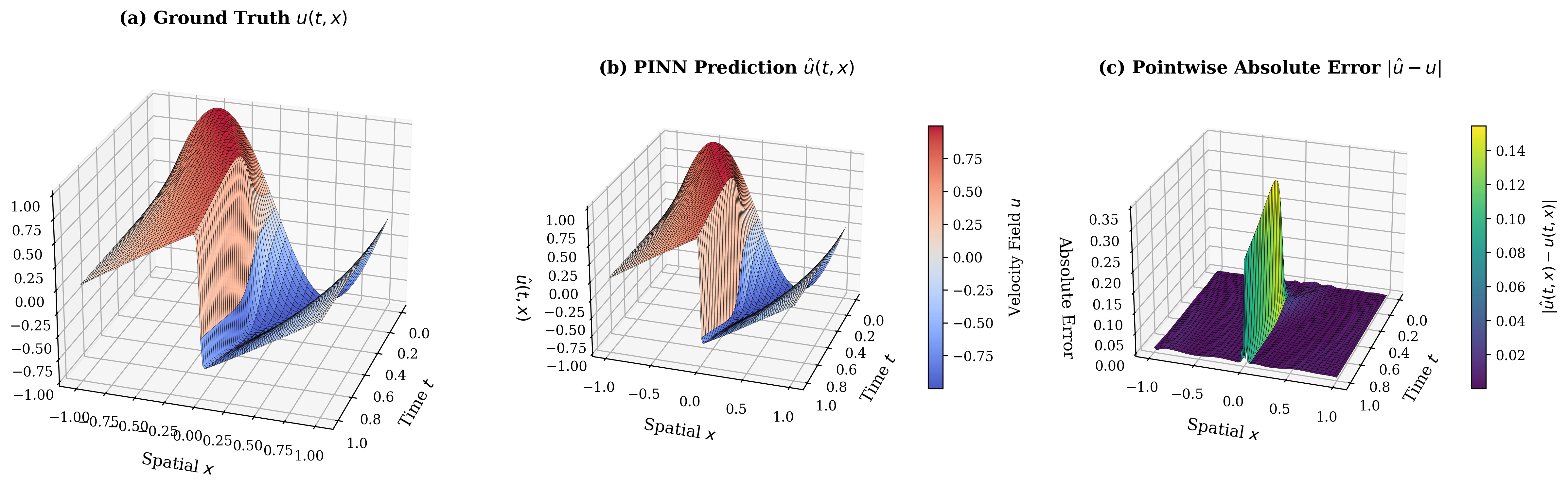}
\caption{3D comparison for the 1D viscous Burgers' velocity field across the in-distribution space-time domain $(t, x) \in [0, 1] \times [-1, 1]$: \textbf{(a)} high-resolution finite-difference reference solution $u(t, x)$, \textbf{(b)} PINN base model prediction $\hat{u}(t, x)$ ($L_2 \approx 3.86\%$), and \textbf{(c)} pointwise absolute error $|\hat{u}(t, x) - u(t, x)|$, illustrating that predictive errors are primarily concentrated along the steep internal shock layer at $x \approx 0$.}
\label{fig:burgers_base_pred}
\end{figure}

The viscous Burgers' PINN follows a fully connected architecture (\texttt{FCN}) matching \texttt{burgers\_temporal\_experiment.py}:
\begin{itemize}
    \item \textbf{Input Layer (\texttt{fcs}):} $\text{Linear}(2 \to 20) \to \tanh$, taking spatial and temporal coordinates $(x, t) \in [-1, 1] \times [0, 1]$.
    \item \textbf{Hidden Blocks (\texttt{fch}):} $7$ sequential hidden blocks indexed $0$ through $6$ (\texttt{fch.0.0} to \texttt{fch.6.0}), each consisting of $\text{Linear}(20 \to 20) \to \tanh$ (total $N_{\text{LAYERS}}=8$ layer stages).
    \item \textbf{Output Layer (\texttt{fce}):} $\text{Linear}(20 \to 1)$ projecting to the scalar velocity field prediction $\hat{u}(t, x)$.
    \item \textbf{Parameters:} 3{,}021 total trainable parameters with kinematic viscosity set to $\nu = 0.01/\pi \approx 0.003183$.
\end{itemize}
Figure~\ref{fig:burgers_base_pred} compares the trained PINN base prediction against the high-resolution numerical reference solution in-distribution ($t \in [0, 1], x \in [-1, 1]$), alongside the pointwise absolute error surface displaying localized discrepancy near the steepening shock front ($x \approx 0$).

\begin{figure}[!htbp]
\centering
\includegraphics[width=0.92\linewidth]{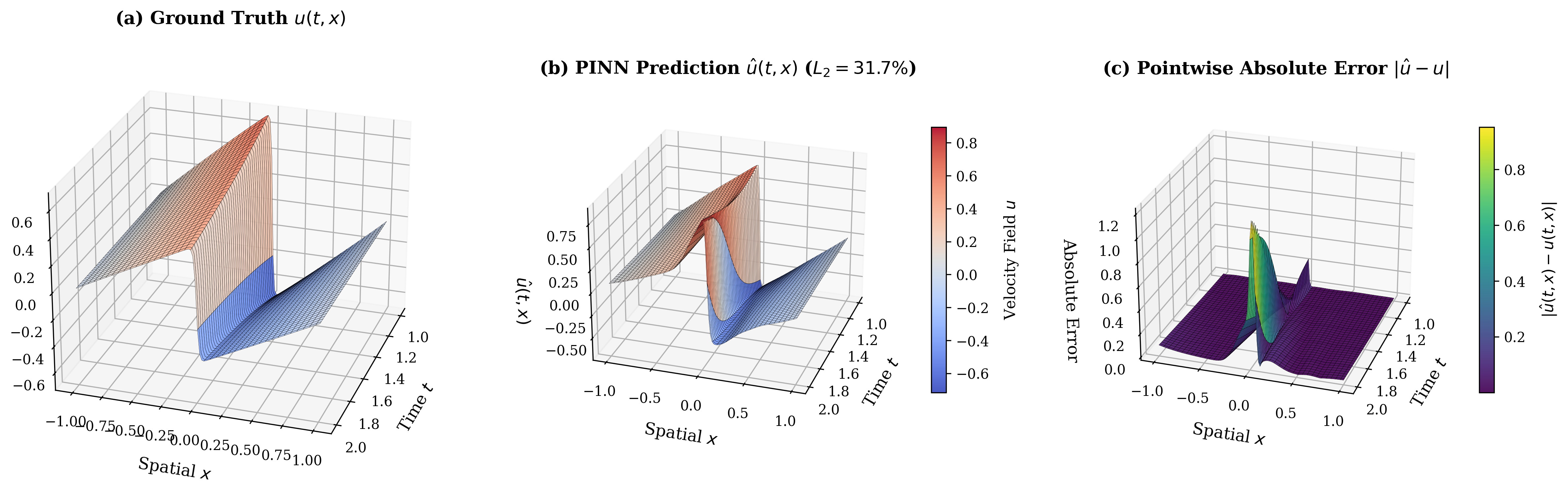}
\caption{3D comparison for the 1D viscous Burgers' velocity field under out-of-distribution temporal extrapolation ($t \in [1, 2], x \in [-1, 1]$): \textbf{(a)} high-resolution reference solution $u(t, x)$ continuing smooth viscous dissipation, \textbf{(b)} PINN model prediction $\hat{u}(t, x)$ showing severe unphysical deformation ($L_2 \approx 31.75\%$), and \textbf{(c)} pointwise absolute error $|\hat{u}(t, x) - u(t, x)|$, highlighting large localized error growth along the extrapolated shock front.}
\label{fig:burgers_ood_pred}
\end{figure}

Under temporal extrapolation ($t \in [1, 2]$), model prediction diverges substantially from the governing physics (Figure~\ref{fig:burgers_ood_pred}), with relative $L_2$ error jumping to $31.75\%$ and maximum pointwise absolute error exceeding $1.22$. While the true solution exhibits gradual viscous dissipation, the unconstrained PINN exhibits severe unphysical distortion and trough sagging near $x \approx 0$.

\clearpage
\section{Technical Appendix: Damped Harmonic Oscillator Benchmark}
\label{sec:appendix_dho}

\subsection{Experimental Setup and Baseline Results}
We evaluated GIIT on the 1D damped harmonic oscillator ($m\ddot{x}+c\dot{x}+kx=0$ with $m=1, c=4, k=400$), modeled with a fully connected PINN ($\text{Linear}(1\to 32)\to\tanh\to\text{Linear}(32\to 32)\to\tanh\to\text{Linear}(32\to 32)\to\tanh\to\text{Linear}(32\to 1)$, 2{,}209 parameters), trained with a combined data-fit and physics-residual loss. The ID regime covers $t \in [0, 1]$ (500 uniform collocation points), while the OOD transfer regime tests temporal extrapolation over $t \in [1, 2]$ (500 collocation points) referenced against the analytical damped trajectory. Note that the reported physics residual ($\text{mean}(R^2)$) measures the raw, unnormalized squared ODE operator norm $(m\ddot{x}+c\dot{x}+kx)^2$; because $k=400$, nominal values of $kx$ and $\ddot{x}$ are on the order of $10^1$ to $10^2$, which naturally leads to an ID residual scale of $\sim 2\times 10^2$ (contrasting with Burgers' equation where normalized fields and $\nu \approx 0.003$ yield smaller numerical residual scales). Under OOD extrapolation, this residual explodes by nearly three orders of magnitude to $1.317\times 10^5$.

\begin{table}[h]
\caption{Oscillator results across In-Distribution ($t\in[0,1]$) and Out-Of-Distribution ($t\in[1,2]$) regimes.}
\label{tab:main_dho}
\centering
\small
\begin{tabular}{lrrl}
\toprule
Regime & Pred.\ $L_2$ error & Physics residual & Layer correspondence \\
\midrule
ID ($t \in [0, 1]$)  & $1.13\%$ & $2.007\times10^{2}$ & Distinct parameter mappings identified \\
OOD ($t \in [1, 2]$) & $2007.34\%$ & $1.317\times10^{5}$ & Mass/damping collapse to shared tensor; spring shifts \\
\bottomrule
\end{tabular}
\end{table}

In-distribution, the model accurately predicts displacement ($L_2 = 1.13\%$) with residual-loss sensitivities $\frac{\partial \mathcal{L}_R}{\partial m} = +1377.28$ and $\frac{\partial \mathcal{L}_R}{\partial c} = -3.84$. Across the 21 perturbation replicates on one trained model, mapping identifies hidden tensors associated with the physical parameters (modal winners \texttt{fch.1.0.weight} and \texttt{fch.0.0.weight}).

\subsection{DHO Out-of-Distribution Transfer Analysis}
\begin{figure}[!htbp]
\centering
\includegraphics[width=0.92\linewidth]{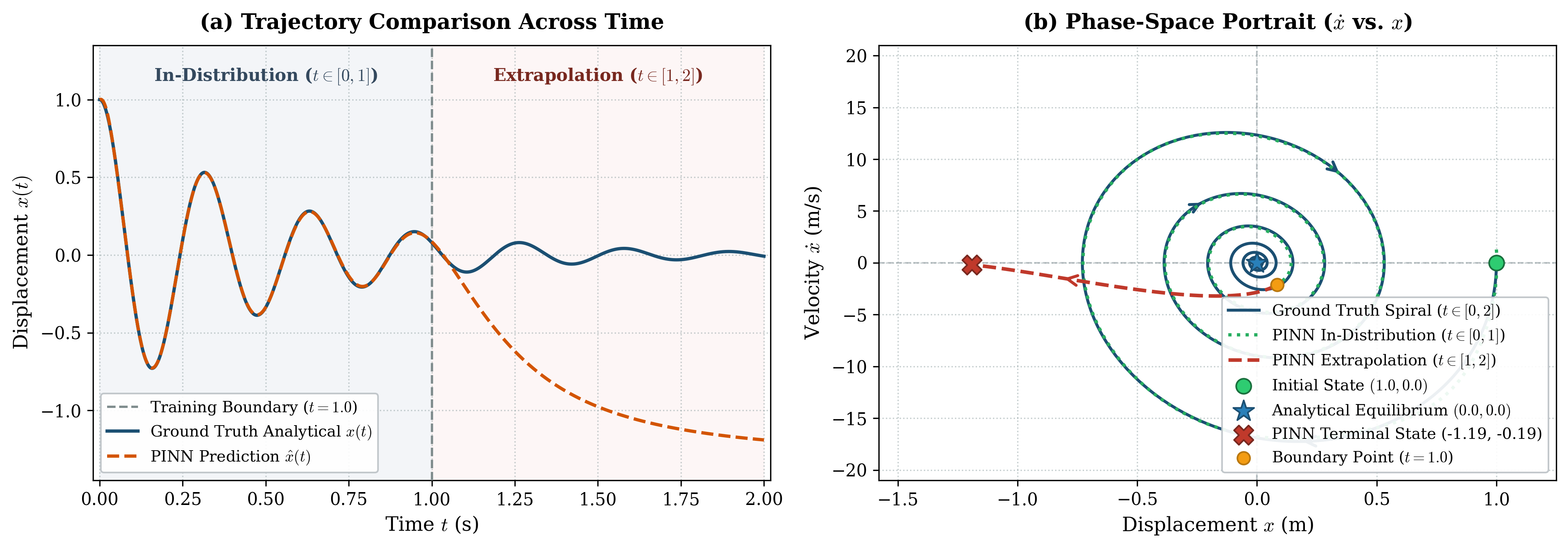}
\caption{Damped harmonic oscillator dynamics in-distribution ($t \in [0, 1]$) and under extrapolation ($t \in [1, 2]$): \textbf{(a)} displacement $x(t)$ comparing ground truth against the PINN, which diverges to a flatline ($\hat{x} \approx -1.19$) after $t=1.0$; \textbf{(b)} phase portrait ($\dot{x}$ vs.\ $x$) showing the true trajectory spiraling into the origin $(0, 0)$, whereas the PINN terminates at $(-1.19, -0.19)$.}
\label{fig:dho_trajectory_phase}
\end{figure}

Evaluating mapping distributions under the OOD regime ($t \in [1, 2]$) reveals significant divergence in residual-loss sensitivity and layer-wise functional correspondence. As shown in Table~\ref{tab:main_dho} and Figure~\ref{fig:dho_trajectory_phase}, while the model achieves low predictive error in-distribution ($L_2 = 1.13\%$), its OOD predictive error explodes ($L_2 = 2007.34\%$), and the physics residual norm increases by nearly three orders of magnitude ($1.317\times10^5$). This catastrophic divergence occurs because the true oscillator trajectory rapidly decays toward equilibrium ($x \to 0, \dot{x} \to 0$), whereas the neural network ceases oscillating near $t \approx 1.0$ and drifts toward an unphysical fixed point ($\hat{x} \approx -1.19, \hat{\dot{x}} \approx 0$). In phase space (Figure~\ref{fig:dho_trajectory_phase}b), while the analytical ground truth forms a decaying spiral converging to the equilibrium origin $(0, 0)$, the PINN trajectory departs from the physical limit cycle at the $t=1.0$ boundary and freezes into a spurious fixed point at $(-1.19, -0.19)$. Dividing the non-decaying prediction error by this vanishing true signal drives the relative $L_2$ error to $2007.34\%$, while the stiff restoring spring force ($k=400$) causes the squared ODE residual $(m\ddot{x}+c\dot{x}+kx)^2$ to explode past $10^5$. In addition, the residual-loss sensitivities shift substantially under extrapolation: $\frac{\partial \mathcal{L}_R}{\partial m}$ shifts from $+1377.28$ to $-1747.92$ and $\frac{\partial \mathcal{L}_R}{\partial c}$ shifts from $-3.84$ to $+579.62$ (detailed in Table~\ref{tab:sensitivities}). Furthermore, re-running auto-mapping in the OOD regime demonstrates that mass and damping mappings collapse to a shared early-layer tensor (\texttt{fch.0.0.weight}), while the spring-constant mapping shifts to a neighboring early layer (\texttt{fch.1.0.weight}).

\begin{table}[htbp]
\caption{Residual-loss Sensitivities ($\partial \mathcal{L}_R/\partial p$).}
\label{tab:sensitivities}
\centering
\small
\begin{tabular}{lcc}
\toprule
Physical Parameter & ID ($t \in [0, 1]$) & OOD ($t \in [1, 2]$) \\
\midrule
Mass ($m$)                  & $+1377.28$ & $-1747.92$ \\
Damping Coeff ($c$)         & $-3.84$    & $+579.62$ \\
Spring Constant ($k$)       & $+111.77$  & $+458.12$ \\
\bottomrule
\end{tabular}
\end{table}

\subsection{Perturbation-Replicate Stability Analysis}
To rigorously evaluate whether the discovered neural-physical mappings are artifacts of a specific random noise draw or represent stable topological properties of the trained PINN, we executed 21 independent perturbation replicates on one trained model ($\text{replicate} \in [0, 20]$ with noise std $\sigma = 0.1$).

Table~\ref{tab:seed_sweep} summarizes the modal (top) mapped neural tensor, the number of replicates where that tensor was selected as the highest-scoring component, and the distribution across candidate layers.

\begin{table}[htbp]
\caption{Neural Component Mapping Distributions across 21 Perturbation Replicates ($\text{replicate} \in [0, 20]$, $\sigma = 0.1$) for both In-Distribution ($t\in[0,1]$) and Out-Of-Distribution ($t\in[1,2]$) evaluation.}
\label{tab:seed_sweep}
\centering
\small
\begin{tabular}{lcccc}
\toprule
Physical Variable & \multicolumn{2}{c}{In-Distribution ($t \in [0, 1]$)} & \multicolumn{2}{c}{Out-Of-Distribution ($t \in [1, 2]$)} \\
\cmidrule(lr){2-3} \cmidrule(lr){4-5}
 & Top Component & Consistency & Top Component & Consistency \\
\midrule
Mass ($m$)                  & \texttt{fch.1.0.weight} & 10/21 (47.6\%) & \texttt{fch.0.0.weight} & 16/21 (76.2\%) \\
Damping Coeff ($c$)         & \texttt{fch.1.0.weight} & 10/21 (47.6\%) & \texttt{fch.0.0.weight} & 16/21 (76.2\%) \\
Spring Constant ($k$)       & \texttt{fch.0.0.weight} & 8/21 (38.1\%)  & \texttt{fch.1.0.weight} & 12/21 (57.1\%) \\
Displacement ($x$)          & \texttt{fch.1.0.weight} & 11/21 (52.4\%) & \texttt{fch.1.0.weight} & 16/21 (76.2\%) \\
Velocity ($\dot{x}$)        & \texttt{fch.0.0.weight} & 17/21 (81.0\%) & \texttt{fch.0.0.weight} & 13/21 (61.9\%) \\
Acceleration ($\ddot{x}$)   & \texttt{fch.0.0.weight} & 16/21 (76.2\%) & \texttt{fch.0.0.weight} & 10/21 (47.6\%) \\
\bottomrule
\end{tabular}
\end{table}

\paragraph{Key Observations from Multi-Replicate Analysis:}
\begin{itemize}
    \item \textbf{Derivative Localization Stability}: Physical derivative states exhibit high consistency in-distribution. Velocity ($\dot{x}$) and Acceleration ($\ddot{x}$) map to the first hidden block (\texttt{fch.0.0.weight}) in 81.0\% and 76.2\% of perturbation replicates, respectively.
    \item \textbf{OOD Correspondence Reorganization}: Under out-of-distribution temporal extrapolation ($t \in [1, 2]$), mass and damping mappings collapse to a shared early-layer tensor (\texttt{fch.0.0.weight} in 76.2\% of replicates), while the spring-constant mapping shifts to a neighboring early layer (\texttt{fch.1.0.weight} in 57.1\% of replicates), demonstrating that the network's internal capacity to separate functional parameter correspondences degrades outside the training envelope.
\end{itemize}

\subsection{Supplementary Material}
Anonymized code, dataset configurations, and model weights are available at \url{https://github.com/adwaitdagitguy/GIIT}.
\newpage
\end{document}